\documentclass[acmtog]{acmart}

\usepackage{amsmath}   
\usepackage{multirow, makecell, array, tabularx, colortbl} 
\usepackage{nicefrac}           
\usepackage{subcaption}         
\usepackage{wrapfig, float}     
\usepackage{algorithm, algorithmicx, algpseudocode} 

\usepackage{diagbox}   
\usepackage{bbding}    
\usepackage{listings}  
\usepackage{comment}   
\usepackage{longtable} 

\AtBeginDocument{%
  }

\setcopyright{none}
\acmDOI{}
\acmISBN{}
\acmPrice{}

\makeatletter
\renewcommand\footnotetextcopyrightpermission[1]{}
\makeatother

\authorsaddresses{}

\setcopyright{acmlicensed}
\copyrightyear{2025}
\acmYear{2025}
\setcopyright{acmlicensed}\acmConference[SA Conference Papers '25]{SIGGRAPH Asia 2025 Conference Papers}{December 15--18, 2025}{Hong Kong, Hong Kong}
\acmBooktitle{SIGGRAPH Asia 2025 Conference Papers (SA Conference Papers '25), December 15--18, 2025, Hong Kong, Hong Kong}
\acmDOI{10.1145/3757377.3763944}
\acmISBN{979-8-4007-2137-3/2025/12}



\begin{document}

\title{AGSwap: Overcoming Category Boundaries in Object Fusion via Adaptive Group Swapping}

\author{Zedong Zhang}
\affiliation{%
  \institution{Nanjing University of Science and Technology}
  \city{Nanjing}
  \country{CHINA}
}
\email{zandyz@njust.edu.cn}

\author{Ying Tai}
\affiliation{%
  \institution{Nanjing University}
  \city{Nanjing}
  \country{CHINA}}
\email{yingtai@nju.edu.cn}

\author{Jianjun Qian}
\affiliation{%
  \institution{Nanjing University of Science and Technology}
  \city{Nanjing}
  \country{CHINA}}
\email{csjqian@njust.edu.cn}

\author{Jian Yang}
\affiliation{%
  \institution{Nanjing University of Science and Technology}
  \city{Nanjing}
  \country{CHINA}}
\email{csjyang@njust.edu.cn}

\author{Jun Li}
\authornote{Corresponding author.}
\affiliation{%
  \institution{Nanjing University of Science and Technology}
  \city{Nanjing}
  \country{CHINA}}
\email{junli@njust.edu.cn}

\renewcommand{\shortauthors}{Zhang et al.}

\begin{abstract}
Fusing cross-category objects to a single coherent object has gained increasing attention in text-to-image (T2I) generation due to its broad applications in virtual reality, digital media, film, and gaming. However, existing methods often produce biased, visually chaotic, or semantically inconsistent results due to overlapping artifacts and poor integration. Moreover, progress in this field has been limited by the absence of a comprehensive benchmark dataset. To address these problems, we propose \textbf{Adaptive Group Swapping (AGSwap)}, a simple yet highly effective approach comprising two key components: (1) Group-wise Embedding Swapping, which fuses semantic attributes from different concepts through feature manipulation, and (2) Adaptive Group Updating, a dynamic optimization mechanism guided by a balance evaluation score to ensure coherent synthesis. Additionally, we introduce \textbf{Cross-category Object Fusion (COF)}, a large-scale, hierarchically structured dataset built upon ImageNet-1K and WordNet. COF includes 95 superclasses, each with 10 subclasses, enabling 451,250 unique fusion pairs. Extensive experiments demonstrate that AGSwap outperforms state-of-the-art compositional T2I methods, including GPT-Image-1 using simple and complex prompts.
\href{https://github.com/NJUSTzandyz/AGSwap}{Project Page} 
\end{abstract}


%
%

\begin{teaserfigure}
  \centering
    \includegraphics[width=0.93\linewidth]{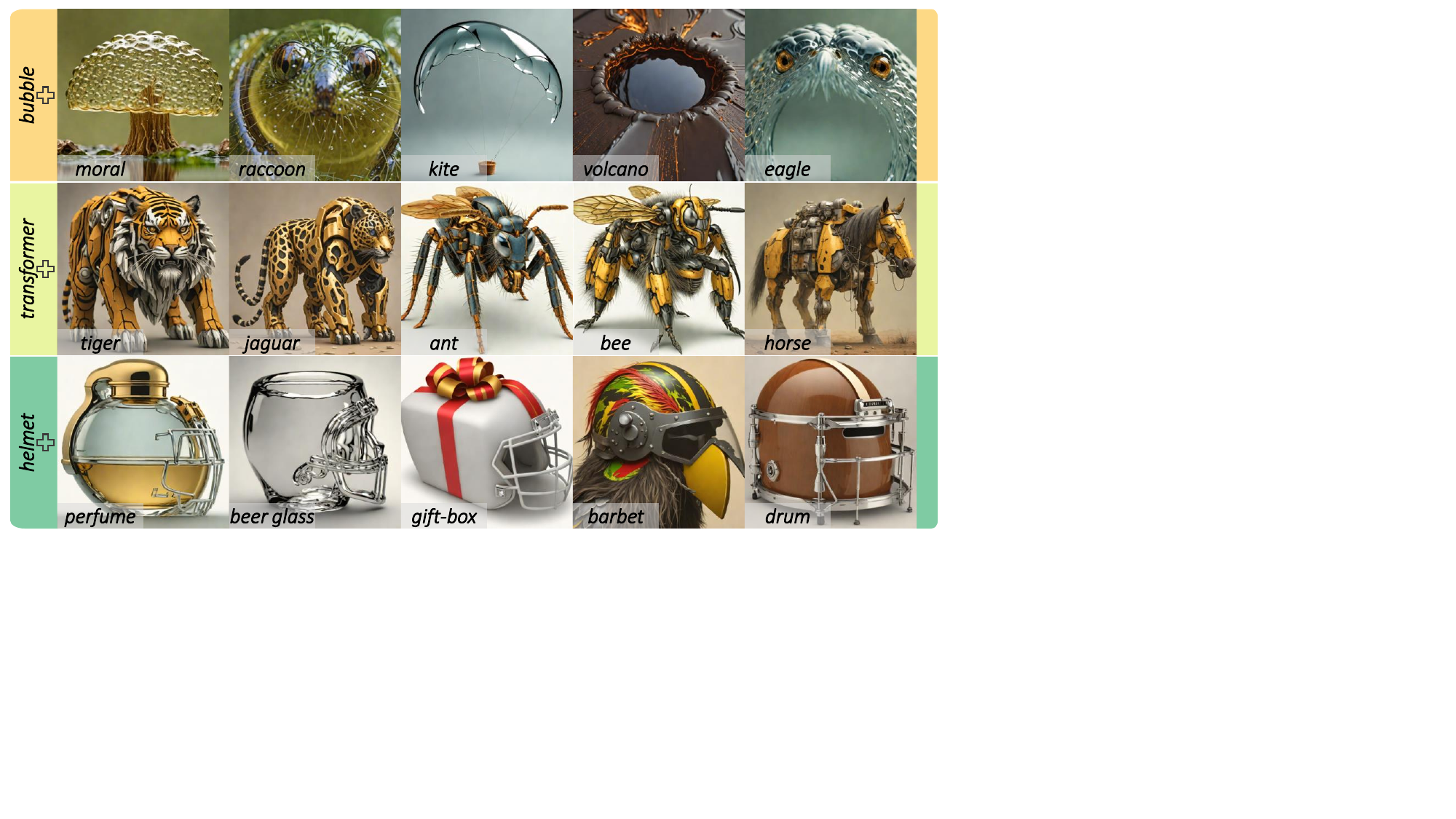}
    \caption{We present a simple yet highly efficient method for seamless object fusion, enabling the creation of novel and visually striking hybrid objects. For instance, as shown in our results, we successfully merge the concept of a \textit{bubble} with diverse concepts such as \textit{moral, raccoon, kite, volcano, and eagle}.\textsuperscript{1}}  
    \label{fig:teaser}
\end{teaserfigure}


\maketitle
\footnotetext[1]{Our method's results, forming the \emph{``Bubble World''} series, were awarded the \textbf{Silver Award} at the NY Digital Awards 2025 (\href{https://nydigitalawards.com/winner-info.php?id=673}{Award Page}, \href{https://www.youtube.com/watch?v=BDMBAXNCVUI}{Video}).}

\section{Introduction}

Text-to-image (T2I) generations \cite{dalle1, ldm, chang2023muse, luo2023lcm, kandinsky} have made significant progress with the development of diffusion models, benefiting from better denoising techniques \cite{ddim, ho2020ddpm, luo2023lcm} and large-scale generative architectures \cite{dalle1, imagen, ldm}. These models excel at producing high-fidelity images that closely match natural language prompts. However, most T2I systems prioritize prompt fidelity and distributional alignment, often struggling to generate imaginative or unconventional outputs—particularly when given minimal, abstract, or compositionally complex prompts \cite{croitoru2023diffusion}. Recent work has increasingly focused on \textit{combinational creativity} \cite{boden1998creativity, boden2004creative} in T2I generation \cite{richardson2024conceptlab,litp2o2024,Xiong2024ATIH,c3,fengcretok2025,conceptcraft,trtst}, where multiple textual concepts are merged to synthesize visually compelling content that diverges from training data in structure, color, or semantics. Building on this, we explore a new combinational T2I approach that generates even more innovative and unexpected results.

\begin{figure}[t]
\centering
\includegraphics[width=0.97\linewidth]{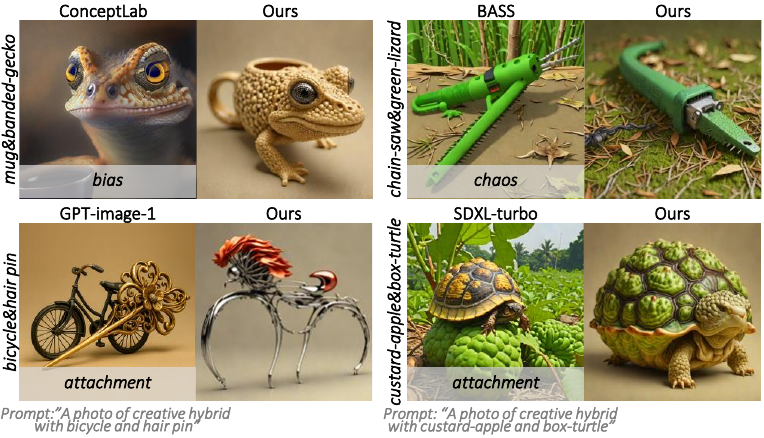}
\caption{Failure examples using ConceptLab \cite{richardson2024conceptlab}, BASS \cite{litp2o2024}, GPT-image-1 \cite{gptimage} and SDXL-turbo \cite{sdxl-turbo}. Our AGSwap produces novel combinational creations successfully.} 
\label{fig:motivation}
\end{figure}

Recent combinational T2I methods can be broadly categorized into two approaches. First, \textbf{prompt-based methods} \cite{gpt4o} is an intuitive method designed prompts that combine multiple text concepts, leveraging models like OpenAI’s GPT-image-1 \cite{gptimage} and StabilityAI's SDXL-turbo \cite{sdxl-turbo} to generate novel outputs. While these models benefit from large-scale training data, they lack fine-grained control over concept fusion, often producing overlapping or incoherent compositions (e.g., failing to properly merge bicycle and hair-pin in Fig. \ref{fig:motivation}, bottom-left). Moreover, crafting effective prompts for diverse and imaginative results remains challenging for non-professional users. 
Second, \textbf{fine-tuning diffusion} models learn new text embeddings or prompt-specific tokens to enhance combinational generations explicitly. For instance, ConceptLab \cite{richardson2024conceptlab} and CreTok \cite{fengcretok2025} use textual inversion \cite{textualinv} to train specialized tokens on curated prompt pairs or abstract concepts. 
However, their high-dimensional token spaces often suffer from sparse inter-class distributions, limiting fusion of semantically distant concepts (e.g., animals and non-animals in Fig. \ref{fig:motivation}, top-left). Alternatively, BASS \cite{litp2o2024} introduces a training-free strategy with random embedding swaps and iterative sampling. Yet, its random swapping vectors lead to inefficient sampling in high-dimensional embedding space, often producing chaotic or biased outputs (Fig. \ref{fig:motivation}, top-right).

To address these limitations, we propose \textbf{Adaptive Group Swapping (AGSwap)}, a simple yet effective method for cross-category combinational T2I generations. Given a category pair \((c_1, c_2)\), we first extract their CLIP embeddings \((E_1, E_2)\), and generate corresponding object images \((I_1, I_2)\) using an edge-cut diffusion model. Our approach begins by computing an initial group swapping vector to blend
the text embeddings \(E_1\) and \(E_2\), producing a mixed embedding \(E_c\). This embedding is then fed into the diffusion model to generate a combined object image \(I_c\). To optimize the fusion, we introduce a visual balance metric based on CLIP scores \cite{clip}, measuring the relative distances between \(d(I_c,I_1)\) and \(d(I_c,I_2)\). This metric serves as feedback to assess fusion bias and dynamically adjust the group swapping vectors. Through iterative refinement—where updated vectors generate improved combinations—AGSwap progressively enhances cross-category fusion, yielding highly coherent and imaginative composite images.

Meanwhile, we construct a \textbf{Cross-category Object Fusion (COF)} dataset since popular datasets like ImageNet \cite{imagenet} often exhibit long-tailed label distributions, with far more animal and plant categories than man-made objects. To mitigate this bias, we reorganize ImageNet into a semantically balanced benchmark. Leveraging WordNet \cite{wordnet}, we group fine-grained categories under shared hypernyms, forming 95 superclasses (10 subclasses each, totaling 950 categories). Unlike ImageNet’s flat structure, our hierarchical design enhances semantic interpretability and enables flexible cross-category combinations (e.g., \textit{animals} + \textit{tools}). This supports $\binom{950}{2}$ possible pairings for object fusion, ideal for combinational generation and large-scale evaluation of cross-category object synthesis. 

\begin{figure*}[!t]
\centering
\includegraphics[width=0.77\linewidth]{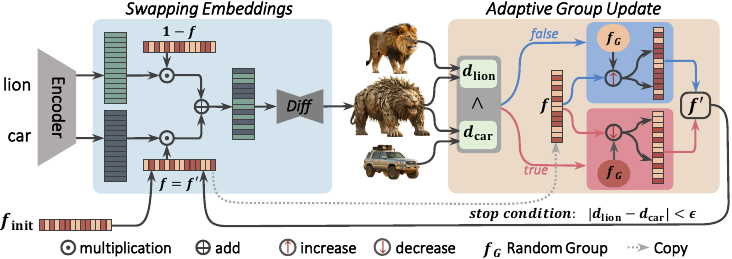}
\caption{Pipeline of our adaptive group swapping, which performs group-embedding swaps and adaptively updates the fusion using a visual balance between the input objects and the blended result.} 
\label{fig:framework}
\end{figure*}

Overall, our contributions are summarized as follows: \textbf{(i)} 
\textit{Adaptive Group Swapping (AGSwap)}: A novel method for cross-category combinational text-to-image generation, enabling flexible and coherent object fusion.
\textbf{(ii)} \textit{Cross-Category Object Fusion Dataset}: A comprehensive benchmark comprising 95 superclasses, each with 10 subclasses, supporting 451,250 unique fusion pairs to facilitate combinational T2I research.
\textbf{(iii)} \textit{State-of-the-Art Performance}: Extensive experiments validate the effectiveness of AGSwap, outperforming recent combinational T2I methods, such as GPT-image-1 \cite{gptimage}, ConceptLab \cite{richardson2024conceptlab}, and BASS \cite{litp2o2024}, in both quality and creativity.

\section{Related Works}

\textbf{Compositional Generation} refers to the ability to generate images that simultaneously contain both input entities, where the inputs can include text or images. Traditionally, it mainly refers to generating an image where two or multi objects appear simultaneously \cite{liu2022compositional, xie2024dymo, li2025moedit, goel2024pair, bao2024separate, c33d}, or where one object is placed naturally within a specified scene or layout \cite{multitwine, noisecollage}. And image editing methods \cite{deutch2024turboedit} can also achieve a certain degree of compositional generation, aiming to transform the content of an image according to the guidance of input text. Specifically, most image editing methods are based on the inversion process of DDPM \cite{friendly} or DDIM \cite{nulltextinv}. For example, PnPInversion \cite{jupnp} reverse the denoising process of DDIM to gradually predict the noise distribution of a given image, and then perform forward denoising from the predicted noise towards the target guided by the input text. In addition, to address the estimation bias in the inversion process, InfEdit \cite{infedit} adopts a consistency equation to directly estimate the original noise based on LCM, effectively improving both the fidelity and editability of the generated images. Meanwhile, benefiting from the trajectory-reversible property of the flow-based schedule \cite{wang2024taming, kulikov2024flowedit} enables image editing by directly inverting the diffusion trajectory, thus avoiding potential deviations.

\textbf{Creative Generation} has attracted increasing attention and is gradually becoming an important branch of diffusion-based image generation. Its primary goal is to generate novel objects, which are surprising and hard to predict in terms of aesthetics or practical value \cite{boden2004creative, boden1998creativity}. Recent methods have explored creative object generation either from a single input or by combining dual inputs.CSFer~\cite{CreativeStyle} proposes a neural permutation network that rearranges features from a single style and, guided by creativity metrics, selects outputs to produce novel, creative stylized images. C3~\cite{c3} enhances the representation of the concept ``creative'' within the denoising process, enabling the generation of creative images from simple text descriptions. In contrast, most existing approaches rely on dual-input combinations to generate new objects that lie between their distributions. BASS~\cite{litp2o2024} blends the word embeddings of multiple prompts, while ConceptLab~\cite{richardson2024conceptlab}, CreTok~\cite{fengcretok2025}, and PartCraft~\cite{ng2024partcraft} adopt textual inversion techniques to introduce new token IDs. ConceptLab and CreTok aim to align the new token with given words, with CreTok further targeting the token ``creative,'' while PartCraft focuses on part-level composition, enabling the fusion of semantic parts from different creatures. Beyond embedding-level fusion, MagicMix~\cite{liew2022magicmix} and ATIH~\cite{Xiong2024ATIH} combine images and text, applying inversion or alignment techniques during the denoising process. Furthermore, Factorized Diffusion~\cite{geng2025factorized} and PTDiffusion~\cite{ptdiffusion} decompose the noise into high- and low-frequency components to control the fusion of an image and a text description. However, for fusion-based methods, although they show considerable fusion ability within the same category, their generalization ability, such as across species (organism-inorganism), is relatively insufficient. BASS provides a certain tendency of generalization ability, but its iterative process brings considerable computational cost, making the search process very time-consuming.

\begin{table*}[!t]
\centering
\scriptsize
\setlength{\tabcolsep}{4pt}
\renewcommand{\arraystretch}{0.97}
\caption{List of All 95 Superclasses in Our Dataset (Sorted by Word Length)}
\resizebox{0.97\linewidth}{!}{%
\begin{tabular}{c|c|c|c|c|c|c|c|c|c|c|c}
\Xhline{1.2pt}
bag & rug & bovid & bottle & machine & clothing & amphibian & pachyderm & echinoderm & coelenterate & other vehicle & sports equipment \\
box & toy & clock & canine & mollusk & covering & appliance & procynoid & plant part & other device & marine mammals & electrical device \\
bus & ball & glass & paddle & padding & edentate & arthropod & reservoir & other craft & other insect & piece of cloth & musical instrument \\
cat & bear & paint & person & primate & footwear & body part & structure & metatherian & invertebrate & sailing vessel & dictyopterous insect \\
dog & bird & shark & rodnet & raptors & penguins & container & viverrine & other plant & other mammal & other equipment & electronic equipment \\
mug & food & snake & weapon & reptile & toiletry & furniture & waterfowl & print media & prototherian & instrumentality & geological formation \\
pot & gear & truck & anapsid & saurian & ungulate & lagomorph & crustacean & woodpeckers & teleost fish & wheeled vehicle & self propelled vehicle \\
ray & tool & beetle & bicycle & aircraft & adornment & musteline & decoration & abstraction & large felines & cypriniform fish &  \\
\Xhline{1.2pt}
\end{tabular}}
\label{tab:superclass_12cols}
\end{table*}

\section{Methodology}

In this section, we introduce an \textbf{Adaptive Group Swapping  (AG-Swap)} method to generate fused object images by combining two distinct category concepts (Fig. \ref{fig:framework}). AGSwap consists of two key components. \textit{Group-wise Embedding Swapping} is a feature manipulation operation that fuses semantic attributes from different concepts. \textit{Adaptive Group Updating} is a dynamic optimization mechanism guided by a balance evaluation score to ensure coherent synthesis. Before presenting our method, we first introduce a simple process of Text-to-Image (T2I) generation.

\textbf{Text-to-Image.} Given a category text $c$ and its corresponding simple prompt $p$: \textit{A photo of <c>}, a T2I model generates an image $I=\mathcal{G}(E)$, where $E=\mathcal{E}(p)\in \mathbb{R}^{h\times w}$ is the text embedding produced by encoder $\mathcal{E}(\cdot)$, and $\mathcal{G}(\cdot)$ maps this embedding to an image $I\in \mathbb{R}^{H\times W}$. In our implementation, we use a pretrained Stable Diffusion model \cite{sdxl-turbo} as our baseline, where $\mathcal{E}(\cdot)$ is the text encoder and $\mathcal{G}(\cdot)$ is the diffusion-based generator. While our framework is model-agnostic and can be adapted to other diffusion models.

\subsection{Group-wise Embedding Swapping}
Given a category pair \((c_1, c_2)\), we construct simple text prompts: $p_1:\text{\textit{A photo of}} <c_1>$ and $ p_2:\text{\textit{A photo of}} <c_2>$. These prompts are fed into the T2I model to generate their corresponding original images: $I_1=\mathcal{G}(E_1)$ and $I_2=\mathcal{G}(E_2)$, where $E_1=\mathcal{E}(p_1)\in \mathbb{R}^{h\times w}$ and $E_2=\mathcal{E}(p_2)\in \mathbb{R}^{h\times w}$. 

\textbf{Embedding Swapping.} Inspired by the swapping mechanism introduced in \cite{litp2o2024}, we present mixing the text embeddings $E_1$ and $E_2$ using a binary exchange vector $f \in \{0,1\}^{w}$ to combine their intrinsic features. The swapping operation is defined as:
\begin{align}
E_f=E_1\times\text{diag}(f)+E_2\times\text{diag}(1-f) \in \mathbb{R}^{h\times w},
\label{eq:swapping}
\end{align}
where $\text{diag}(f)$ denotes the diagonal matrix formed from $f$ and $\times$ represents matrix multiplication. Specifically, the exchange vector is initialized by randomly setting half of its positions to $1$ and the other half to $0$. This swapping embedding $E_f$ is fed into the T2I model to generate an object image $I_f=\mathcal{G}(E_f)$.

\textbf{Group-wise Embedding Swapping.} Building upon the embedding swapping in Eq. \eqref{eq:swapping}, we define an exchange group $G\subset C$, where $C=\{1,2,3,\cdots,w\}$ is the set of column indices. The exchange vector $f$ is modified for indices $i\in G$ by setting $f_i$ to 0 or 1, while keeping other entries unchanged. This yields a modified vector: 
\begin{align}
f'=f_{G}\leftarrow 0\ \text{or}\ 1,
\label{eq:groupswapping}
\end{align}
where $|G|=l$ is the group size. Applying the Eq. \eqref{eq:swapping} with $f'$ produces the modified embedding $E_{f'}$,  which is then passed to the T2I model to generate a composite image $I_{f'}=\mathcal{G}(E_{f'})$.

\subsection{Adaptive Group Updating}
\textbf{Balance Score.} To quantify the bias of the combined image $I_f$ toward $I_1$ versus $I_2$,  we define a balance score $s$ as:
\begin{align}
s = d_1 - d_2, \text{where} \ \ \ d_1=d(I_f, I_1), \ \ \  d_2=d(I_f, I_2),
\label{eq:score}
\end{align}
and $d(\cdot, \cdot)$ measures cosine similarity between CLIP embeddings. The sign of $s$ indicates the directional bias: $s>0$ means that $I_f$ is more aligned with $c_1$ and lacks $c_2$-related content, and  $s<0$ means that $I_f$ favors $c_2$ over $c_1$. The magnitude $|s|$ reflects the degree of bias: large $|s|$ is strong dominance of one category’s features, with minimal representation of the other, and small $|s|$ is a balanced incorporation of both categories, suggesting higher balance. 

\textbf{Adaptive Group Updating.} To ensure the composite image $I_f$ represents an equal interpolation between the two categories, we enforce the constraint $|s|\rightarrow 0$ by proposing an adaptive group updating algorithm. Starting from an initial or previous exchange vector $f^t$, we compute a balance score $s^t$ using Eq. \eqref{eq:swapping} and Eq. \eqref{eq:score}. We then use the positive and negative $s^t$ to refine the adjustment of $E_{f^t}$ toward either $E_1$ or $E_2$ through a group-wise swapping operation in Eq. \eqref{eq:groupswapping}. Specifically, we randomly select a subset of elements in the exchange vector $f^t$ and flip their values (from 0 to 1 or vice versa), obtaining a modified vector $f^{t+1}$. Formally, adaptive group updating is defined as:
\begin{align}
f^{t+1} =
\begin{cases}
f_G^t\leftarrow 1, \ \ \  G\overset{\text{ran}}{\subset} \{i\mid f_i^t=0\}, & \text{if}  \ \ \  s^t>0, \\
f_G^t\leftarrow 0, \ \ \ G\overset{\text{ran}}{\subset} \{i\mid f_i^t=1\}, & \text{if} \ \ \  s^t<0, 
\end{cases},
\label{eq:modifiedvector}
\end{align}
where $|G|=l^t$ denotes the group size at timestep $t$. Using this update rule, the algorithm terminates when the condition $|s^{t+1}|<\epsilon$ is met, where $\epsilon$ is a predefined threshold (in this paper we set $\epsilon=0.01$). We get the final exchange vector $f^*$ is obtained and the modified embedding $E_{f^*}$ is computed using Eq. \eqref{eq:swapping}. This updated embedding is then passed into the T2I model to generate a new image $I_{f^*}=\mathcal{G}(E_{f^*})$, featuring 
 cross-category combined object characteristics.

\textbf{Group Size Selection $l^t$.} We present a size adaptation strategy inspired by gradient descent. Starting from an initial size $l_{\text{init}}$, the group size $l^t$ is progressively reduced by $\Delta l$ if the update direction oscillates. If the size reaches $l_{\text{min}}$ without convergence, it is reset to $l_{\text{init}}$. Oscillation is detected as a sign flip in $s^t$ between consecutive iterations, and the size is adjusted after four such flips. Through empirical analysis, we set $l_{\text{init}}=10$ to balance convergence speed and stability. 

\begin{table*}[t] 
\setlength{\tabcolsep}{3pt}
\renewcommand{\arraystretch}{0.97}
\centering
\caption{Quantitative Comparison using CLIP, VQA, ChatGPT-4o.}
\resizebox{0.77\linewidth}{!}{%
\begin{tabular}{c|cc|c|cccc}
\Xhline{1.2pt}
\multirow{2}{*}{Model} & \multicolumn{2}{c|}{CLIP Score \cite{clip}} & \multicolumn{1}{c|}{VQA Score \cite{vqascore}}   & \multicolumn{4}{c}{ChatGPT-4o \cite{gpt4o}}\\ 
\cline{2-8}
   &  avg. sim.$ \uparrow $   & balance$ \downarrow $ & text-sim.$ \uparrow $  & surprise$ \uparrow $ & value$ \uparrow $   & novelty$ \uparrow $ & overall$ \uparrow $\\
\hline
\rowcolor{green!10}  Ours  & \textcolor{blue!70}{0.56} & \textcolor{red!70}{0.01} & \textcolor{blue!70}{0.58}  & \textcolor{red!70}{7.3} & \textcolor{red!70}{7.4} & \textcolor{red!70}{7.9}  & \textcolor{red!70}{7.6} \\
BASS{\tiny (ECCV'24)}    & 0.52 & 0.12 & 0.53  & 7.1 & 6.0 & \textcolor{blue!70}{7.2}  & 6.5 \\
ConceptLab{\tiny(TOG'24) }          & 0.38 & \textcolor{blue!70}{0.11} & 0.51  & \textcolor{blue!70}{7.2} & 5.8 & 7.1  & 6.7 \\
SDXL-turbo{\tiny (StabilityAI'23)}         & 0.48 & 0.17 & 0.47  & 6.5  & \textcolor{blue!70}{7.2} & 6.8  & 6.2 \\
GPT-image-1{\tiny(OpenAI'25)}     & \textcolor{red!70}{0.61} & \textcolor{blue!70}{0.11} & \textcolor{red!70}{0.62}  & 7.0  & \textcolor{red!70}{7.4} & \textcolor{blue!70}{7.2}  & \textcolor{blue!70}{7.1} \\
\Xhline{1.2pt}
\end{tabular}
}
\label{table:performance_comparison}
\end{table*}

\section{Cross-category Object Fusion (COF) Dataset}

To comprehensively evaluate cross-category creative fusion, we construct a large-scale dataset named the Cross-category Object Fusion (COF) Dataset. Part of its categories originate from the textual labels of ImageNet-1k \cite{imagenet}, and are reorganized and supplemented using WordNet \cite{wordnet}. .

\textbf{Motivation.} When using existing popular datasets (e.g., CangJie \cite{fengcretok2025}, ImageNet-1K \cite{imagenet}, and CIFAR-100 \cite{cifar100}) for cross-category combinational T2I generation, several limitations arise. First, CangJie (constructed by CreTok \cite{fengcretok2025}) contains fewer than 100 textual categories, primarily limited to animals and plants, which restricts the diversity of possible textual pair combinations. Second, ImageNet-1K, while rich in labeled categories, suffers from a pronounced long-tail distribution. For instance, it includes around 20 \textit{car}-related categories, while \textit{dog}-related categories vastly outnumber others. This imbalance skews the evaluation of cross-category fusion performance. Additionally, ImageNet’s flat categorical hierarchy—compr-ising only basic-level categories—further limits its suitability for hierarchical cross-category tasks. Third, CIFAR-100 provides well-defined superclass-subclass relationships but lacks sufficient scale and diversity, with far fewer categories and a limited coverage of human-made objects. Given these constraints, therefore, constructing a hierarchically structured dataset with balanced category distributions and broader semantic coverage is essential for advancing cross-category combinational T2I generation.

\textbf{Dataset Construction.}
To this end, we construct our COF dataset based on the textual labels of ImageNet-1k \cite{imagenet}, reorganized using hierarchical semantic relationships from WordNet \cite{wordnet}. The constructed process is defined as the following three steps:

\textit{Superclass Candidate Set.} ImageNet-1K has a category set $C=\{c_i\}_{i=1}^{1,000}$ and each category $c_i$ is treated as a subclass. For every $c_i$, we sample a hypernym path $p_i = \{s_i^j\}_{j=0}^{k_i}$ from WordNet, extending from $c_i(s_i^0 = c_i)$ up to the root hypernym $[\textit{object}](s_i^{k_i} = [\text{\textit{object}}])$, where $k_i$ denotes the path length of the category $c_i$. The superclass candidate set is then defined as:
\begin{align}
S_{\text{can}}=\underset{i}{\bigcup} \hat{p}_i, \ \ \ \text{where} \ \ \ \hat{p}_i= \{s_i^j\mid S_i^j\cap C \neq \emptyset\}_{j=1}^{k_i-1},
\label{eq:scan}
\end{align}
where $S_i^j$ represents a set of subordinated words of node $s_i^j$, and $|S_{\text{can}}|= 485$ denotes the candidate number of superclasses.

\textit{Manual Superclass Selection.} The candidate set $S_{\text{can}}$
generated through WordNet’s hierarchy, presents limitations due to insufficient consideration of visual similarities among animal categories. For instance, the hypernym \textit{canine} includes hyponyms like \textit{fang}, \textit{dog}, \textit{fox}, \textit{hyena}, \textit{jackal}, \textit{wild\_dog}, and \textit{wolf}.
While some (e.g.,\textit{dog} and \textit{wild\_dog}) are visually similar, others differ significantly. To ensure visual coherence, we manually refine candidate superclasses by retaining only the most representative category (e.g., keeping \textit{dog} over \textit{wild\_dog}). This results in a semantically meaningful and visually consistent superclass set. After manual curation, we get a manual set $S_{\text{man}}$ with 95 superclasses in Table \ref{tab:superclass_12cols}. The deleted superclasses are shown in Appendix B.

\textit{Subclass Balancing and Augmentation.} After finalizing the superclasses, we balance and augment their subclasses. First, we map subclasses to their corresponding superclasses using hypernym relationships. For superclasses with more than 10 subclasses, we randomly reduce the number to 10. Conversely, if a superclass has fewer than 10 subclasses, we expand the set by identifying hyponyms from WordNet. All subclasses are shown in Appendix B.

By following these procedures, we finalize the COF dataset, which consists of 95 superclasses with 10 subclasses each, resulting in 950 distinct textual categories. This hierarchical structure enables the generation of $\binom{950}{2}=451,250$ unique textual pairs.

\section{Experiments}

\subsection{Experiment Settings}

\textbf{Dataset.}
Although we constructed a large-scale dataset containing 950 categories, due to the high computational cost of BASS and ConceptLab, as well as the pricing constraints of GPT-image-1, we randomly selected one subclass from each superclass to form a smaller subset named COF-tiny. This subset includes 95 textual categories, resulting in a total of 4465 possible text pairs for comparison. The COF-tiny dataset is provided in Appendix B.

\textbf{Evaluation Metrics.} We assess the creativity and semantic balance of our results using CLIP, VQA, and ChatGPT-4o. \textbf{CLIP score:} We compute two metrics using a pretrained CLIP model \cite{clip}: \textit{average similarity} $\frac{1}{2} \left(\cos(I_f, I_1) + \cos(I_f, I_2)\right)$ reflects the overall semantic relevance of the fused image to the source concepts-higher values indicate better alignment, and absolute difference $\left|\cos(I_f, I_1) - \cos(I_f, I_2)\right|$, evaluates fusion fairness—lower values indicate more balanced blending. For fairness, $I_1$ and $I_2$ are generated independently from the same prompts $(p_1, p_2)$, avoiding reference images used in fusion. \textbf{ VQA score:} For each category pair \(c_1,c_2\), we assess prompt-image alignment using a CLIP-based VQA model \cite{vqascore} to compute the similarity between the generated image and the canonical prompt: \textit{A realistic photo of creative hybrid with $c_1$ and $c_2$}. \textbf{ChatGPT-4o:} We leverage ChatGPT-4o \cite{gpt4o} to evaluate creativity based on three criteria \cite{boden2004creative}: \textit{\textbf{Surprise (1--10)}}: \textit{Evaluate whether the object combination or fusion in the image is easily imaginable or visually impressive.} \textit{\textbf{Value (1--10)}}: \textit{Evaluate whether the image conveys realism, focusing on texture, depth, lighting, and details that make the object appear believable.} \textit{\textbf{Novelty (1--10)}}: \textit{Evaluate whether the objects or composition present new or rarely seen concepts.} \textit{\textbf{Overall (1--10)}}: \textit{Provide an overall score based on the three aspects above.} 

\textbf{Experimental Details.}
We use SDXL-Turbo \cite{sdxl-turbo} as our baseline T2I model, which incorporates two text encoders: a CLIP text encoder (ViT-G/14) \cite{clip} as the base encoder and a T5-XXL encoder \cite{T5} as the pooled text encoder. For embedding swapping, we exclusively use the ViT-G/14's embeddings, while the T5 encoder’s pooled embeddings undergo linear interpolation with a fixed ratio of 0.5—a choice we found to have negligible impact on results. For CLIP-based evaluation, we use a ViT-bigG/14-backed CLIP model as our feature extractor. To ensure precise image-content similarity, we extract the main subject from each image using RMBG-2.0 \cite{rmbg} as our segmentation model. In Group Size Selection, we set $\Delta l$ to 2.

\subsection{Main Results}
We compare our method with several representative baselines. For combinational T2I methods, we compare with BASS \cite{litp2o2024} and ConceptLab \cite{richardson2024conceptlab}, as they share similar input settings and generation objectives. For prompt-based methods, we benchmark against SDXL-Turbo model \cite{sdxl-turbo} and GPT-image-1 \cite{gptimage}. Thus, we have the following comparisons.

\subsubsection{Quantitative  Comparison}
Table \ref{table:performance_comparison} reports the quantitative results, with the following key observations. First, our method achieves the best balance score (0.01), indicating the most balanced alignment with minimal bias toward either input. This underscores its ability to maintain symmetrical fusion without favoring one concept. In contrast, BASS and SDXL-Turbo show more imbalance, emphasizing a single input and compromising fusion coherence.

Seond, while GPT-image-1 slightly outperforms our method in average CLIP score by a narrow margin (0.05), this is largely attributed to its tendency to produce object juxtapositions rather than true semantic fusion. A similar trend appears in the VQA Score: GPT-image-1 ranks highest by clearly retaining both objects but fails to achieve deep semantic fusion. Our method, ranking second, delivers more coherent and semantically rich synthesis. BASS and ConceptLab show moderate results but often suffer from visual inconsistency or semantic ambiguity.

Third, in GPT-4o-based evaluation, our method outperforms all baselines across all criteria. Notably, it achieves a 0.7-point lead in \textit{novelty} over the second-best methods (BASS and GPT-image-1), highlighting its superior ability to generate novel concepts. Additionally, our method scores highest in \textit{value} and \textit{surprise}, confirming its capacity to produce images that are not only creative and visually compelling but also practically meaningful. ConceptLab lags in these metrics, likely due to its constrained semantic range.


\subsubsection{Visual Comparison}
\begin{figure*}[t]
\centering
\includegraphics[width=0.97\linewidth]{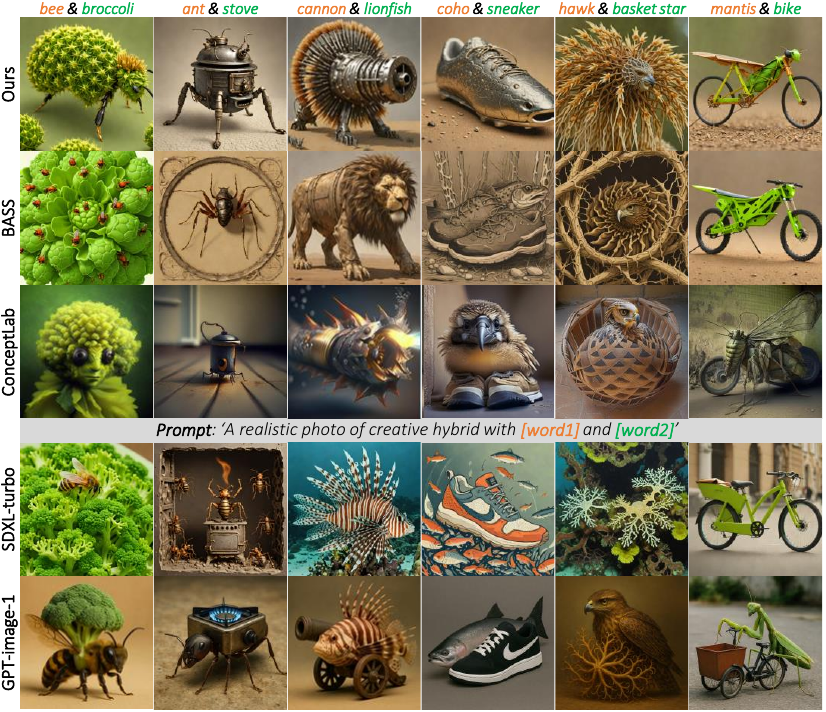}
\caption{Comparison with combinational T2I methods (e.g., BASS \cite{litp2o2024} and ConceptLab \cite{richardson2024conceptlab}) and prompt-based methods (e.g., SDXL-Turbo \cite{sdxl-turbo} and GPT-image-1 \cite{gptimage}). Compared to them, our approach demonstrates superior object blending.} 
\label{fig:res_simple}
\end{figure*}

\textit{\textbf{Comparison with the combinational T2I methods.}}
The first three rows of Fig. \ref{fig:res_simple} compare our method with ConceptLab \cite{richardson2024conceptlab} and BASS \cite{litp2o2024}.
Our method consistently outperforms combinational T2I baselines, which maintains semantic coherence and balanced feature integration even with semantically distant input pairs. For instance, given the pair "\textit{ant-stove}", our method synthesizes a novel stove with ant-like legs and antennae, forming a unified, product-like design. In contrast, BASS generates a spider-like image that deviates from both inputs, while ConceptLab produces a visually plausible object lacking identifiable features from either concept.

Similarly, for "\textit{coho-sneaker}", our model effectively embeds realistic coho textures into the sneaker design. BASS, constrained by fixed vector proportions, generates a generic sneaker with poorly integrated fish attributes. ConceptLab fails to integrate recognizable fish elements, reflecting token-based training’s limitations on semantically distant concepts. In both cases, our method achieves more meaningful fusion and superior visual-semantic alignment.

\noindent \textit{\textbf{Comparison with the prompt-based methods.}}
The last two rows of Fig.~\ref{fig:res_simple} compare our method with prompt-based generative models, specifically SDXL-Turbo and GPT-image-1. For these models, we used the prompt: "\textit{A realistic photo of creative hybrid with [word1] and [word2].}" Our method also surpasses the prompt-based approaches, which often fail to produce true hybrids.
For "\textit{mantis-bike}", SDXL-Turbo typically generates a standard bike, omitting mantis features entirely. GPT-image-1 includes both concepts but merely juxtaposes them—evident in its "\textit{coho-sneaker}" output, where the fish appears beside the shoe rather than being integrated. These limitations underscore the challenges of prompt-only generation, which tends to result in either concept omission or superficial co-location. In contrast, our model explicitly fuses both inputs into a novel, coherent entity with seamless semantic integration.

\begin{figure}[t]
    \centering
    \includegraphics[width=0.97\linewidth]{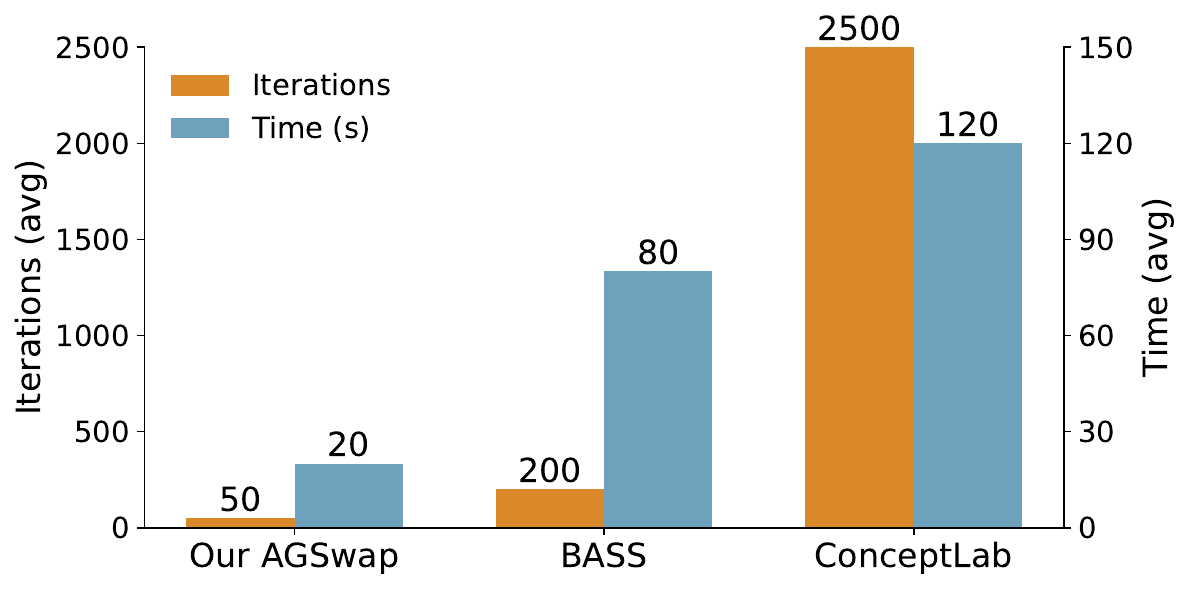}
    \caption{Time cost comparison of different combinational methods.} 
    \label{fig:time}
\end{figure}

\subsubsection{Time Cost~\mbox{Comparison}}

Fig. \ref{fig:time} compares average iterations and runtime for our AGSwap, BASS, and ConceptLab. On an NVIDIA GeForce RTX 4090, AGSwap achieves the best efficiency—$\sim$4$\times$ faster than BASS and $\sim$6$\times$ faster than ConceptLab—while using $\sim$4$\times$ and $50\times$ fewer iterations—making it the most efficient method.

\subsubsection{Complex prompt comparison}
To further evaluate our method’s capability in complex-prompt generation, we conduct a comparative study using SDXL-Turbo and GPT-4o with two distinct prompt strategies. 1) GPT-4o-Generated Prompts: We instruct GPT-4o with: “\textit{Generate a descriptive prompt (under 40 words) that guides a text-to-image model to create a creative hybrid of [word1] and [word2].}” These prompts focus on compositional characteristics but often produce fragmented or disjointed outputs. 2) Image-Derived Prompts: We extract prompts from our own generated images by instructing GPT-4o:
“\textit{Describe this hybrid of [word1] and [word2] in detail.}”
These prompts capture richer attributes like material, texture, and spatial relationships, leading to more cohesive fusions. This dual approach ensures a rigorous and fair evaluation of compositional fidelity and visual richness.

As shown in Fig. \ref{fig:res_complex}, GPT-4o’s original prompts tend to describe objects as separate components—for example, the “\textit{great grey owl-drone}” prompt results in disjointed compositions. In contrast, our image-derived prompts yield more nuanced descriptions (e.g., “\textit{gol-den jointed limbs and a chrome-like abdomen}” for “\textit{bubble-black widow}”), enhancing SDXL-Turbo’s ability to generate well-fused hybrids. Notably, SDXL-Turbo struggles with GPT-4o’s unstructured prompts, often failing to clearly render objects (e.g., titi in “\textit{titi-loggerhead}”) or merely juxtaposing elements (e.g., “\textit{bubble-black widow}”). However, with our refined prompts—which incorporate richer visual cues, spatial logic, and material details—SDXL-Turbo produces significantly more cohesive and imaginative fusions.

\begin{figure}[t]
    \centering
    \includegraphics[width=0.92\linewidth]{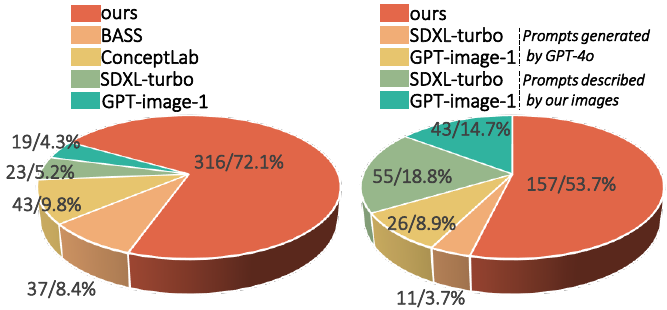}
    \caption{User Study of combinational methods comparison results (left) and complex prompts results (right).} 
    \label{fig:usr_stu}
\end{figure}

\subsection{User Study}
\begin{figure*}[t]
\centering
\includegraphics[width=0.97\linewidth]{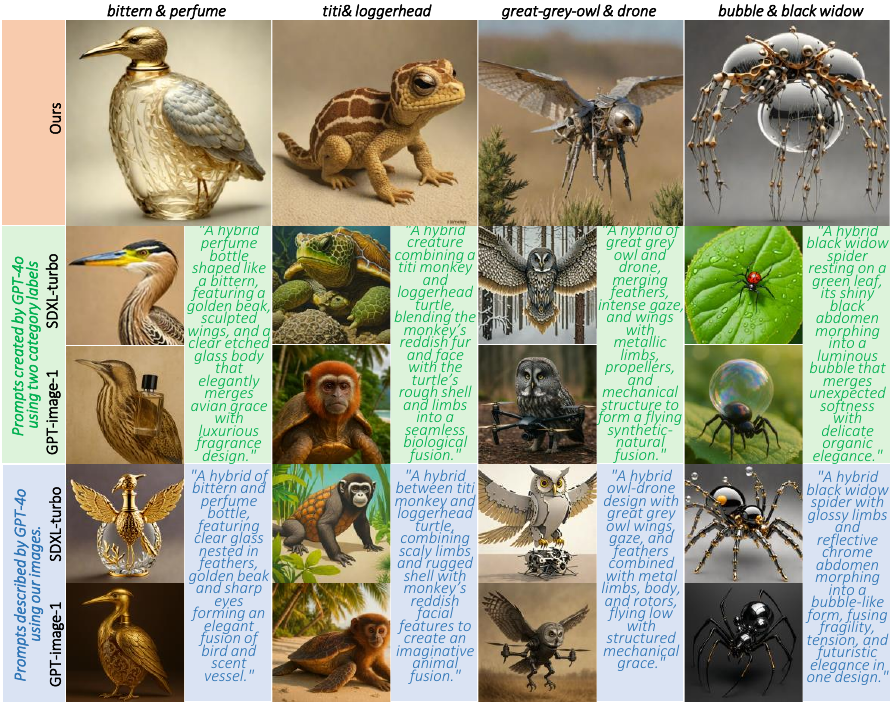}
\caption{\textbf{Comparison with prompt engineering}. We evaluate our method against SDXL-Turbo and GPT-image-1 , using GPT-4o to generate prompts. \textit{Top (Our AGSwap)}: Successfully fuses objects into a coherent result. 
\textit{Middle}: Prompts are generated by GPT-4o from two category labels, leading to less precise compositions.
\textit{Bottom}: Prompts are derived by GPT-4o from our synthesized images, yet still struggle with seamless fusion. } 
\label{fig:res_complex}
\end{figure*}

We conducted user studies to compare our method with combinational approaches as well as with methods using complex prompt texts. Each participant evaluated the results of six groups of combinational comparisons in Fig. \ref{fig:res_simple}, and four groups of complex prompt comparisons in Fig. \ref{fig:res_complex}. This resulted in 438 votes for the combinational comparison and 292 votes for the complex prompt comparison, collected from 73 users. The results for the combinational methods are shown in the left side of Fig. \ref{fig:usr_stu}. Our method received 72.1\% of the votes, significantly outperforming other methods, demonstrating a clear advantage in perceived creativity. In contrast, BASS and ConceptLab, which are similar in approach to ours, received only 3.7\% and 8.9\% of the votes, respectively. Notably, although GPT-image-1 appears visually better than SDXL-turbo in the figure, it performed poorly in the user study, likely due to its tendency to generate simple concatenations, which reduces perceived creativity.

From the right side of Fig. \ref{fig:usr_stu}, our method also achieved the highest performance in the complex prompt comparison, obtaining 53.7\% of the votes. Both generative methods showed noticeable improvement when using our images as the basis for prompt construction. Interestingly, when SDXL-turbo used our image-based prompts, it outperformed GPT-image-1—reversing an initial 5.2\% deficit (3.7\% vs. 8.9\%) to a 4.1\% lead (18.8\% vs. 14.7\%). This demonstrates the potential of our method to enhance prompt engineering, enabling less capable generative models to produce more creative outputs. More details of user study are shown in appendix C

\begin{figure}[t]
\centering
\includegraphics[width=0.97\linewidth]{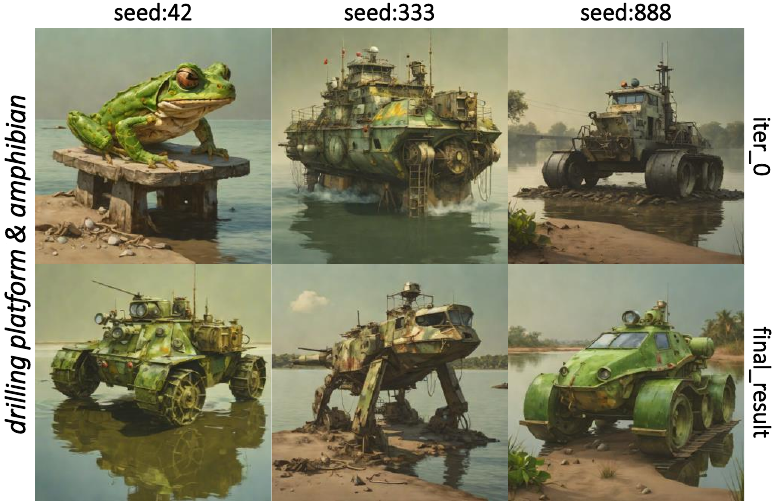}
\caption{Initial and final generations under different random seeds.}
\label{fig:sup_cohe}
\end{figure}

\subsection{Why AGSwap can generate coherent objects}
\begin{figure*}[t]
\centering
\includegraphics[width=0.92\linewidth]{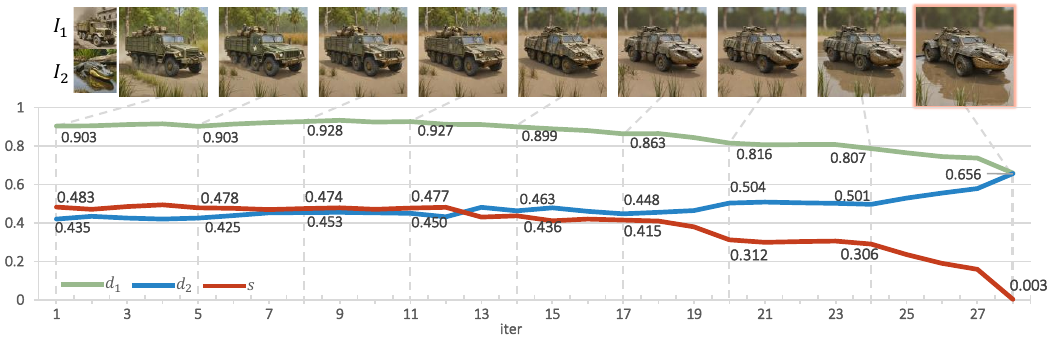}
\caption{The iterative process of our AGSwap method for fusing \textit{half-truck} and \textit{alligator}. Here, $d_1(I_f,I_1)$, $d_2(I_f,I_2)$ and $s=d_1-d_2$ are defined in Eq. \eqref{eq:score}. } 
\label{fig:iter}
\end{figure*}

Our AGSwap method produces coherent objects—rather than two co-occurring objects—for three reasons.
\textit{First—Single-conditioning prior}. As shown in Fig. \ref{fig:sup_cohe}, the top row (iter\_0) displays initial generations under different seeds; the results are already predominantly single objects near the balance region, not simple side-by-side placements. This arises because Group-wise Embedding Swapping with a binary change vector \(f\) constructs one mixed conditioning
\(E_f = E_1\,\mathrm{diag}(f) + E_2\,\mathrm{diag}(1-f)\)—not two parallel conditionings—so SDXL’s cross-attention repeatedly queries a single token sequence.
\textit{Second—Small-step group updates around one embedding}. In Fig. \ref{fig:sup_cohe} the bottom row (final\_result) and Fig. \ref{fig:iter}, our group-based updating strategy reduces the balance score by flipping a small number of exchange groups while staying around the same mixed embedding; changes predominantly add the missing cues from the input pair (shape or texture) and are insufficient to create a new independent subject.
\textit{Third—Subject-aware balance penalizes juxtaposition}. We compute the image--image balance score on the foreground subject (RMBG-2.0). Whenever two separate objects appear, the foreground typically covers only one, so \(\lvert s\rvert\) remains large; the update rule is therefore forced to inject exchange groups of the ``missing'' concept into the same subject until \(\lvert s\rvert\) decreases. This instance-level constraint naturally penalizes juxtaposition and encourages key cues from both sides to converge on one foreground entity, favoring the emergence of a single fused object.

\subsection{Parameter Analysis and Ablation Study.}

\textbf{Parameter Analysis.} \textit{Iterative process.} Fig. \ref{fig:iter} illustrates the iterative fusion process of AGSwap, combining the \textit{half-truck} and \textit{alligator} concepts. As the fusion progresses, the resulting object $I_f$ gradually incorporates more alligator features while reducing half-truck characteristics, ultimately converging toward a balanced representation.

\begin{table}[t]
\setlength{\tabcolsep}{5pt}
\renewcommand{\arraystretch}{1.1}
\centering
\caption{Parameter analysis of the $l_{\text{init}}$. }
\resizebox{0.9\linewidth}{!}{%
\begin{tabular}{c|ccccc}
\Xhline{1.2pt}
$l_{\text{init}}$      & 5 & \textbf{10} &  {15} & {20} & {25} \\
\hline
$iter_{avg}/\text{time}_{avg}$     & 45/19s & 50/20s & 56/21s & 58/21s & 70/30s \\
$CLIP sim._{avg}$     & 0.51 & 0.56 &  0.56 & 0.55 & 0.58 \\
\Xhline{1.2pt}
\end{tabular}}
\label{table:parameteranalysis}
\end{table}

\textit{Group size.} The key parameter in our AGSwap method is $l_{\text{init}}$, which determines the number of iterations needed to find the optimal swapping vector. As shown in Table \ref{table:parameteranalysis}, we evaluated this parameter across a range of values (5 to 25) by measuring the average iterations, search time, and mean CLIP similarity. Based on these experiments, we set the default $l_{\text{init}}$ to 10, as it strikes the best balance between convergence speed and CLIP similarity—achieving the second-highest performance in both metrics.

\begin{figure}[t]
\centering
\includegraphics[width=\linewidth]{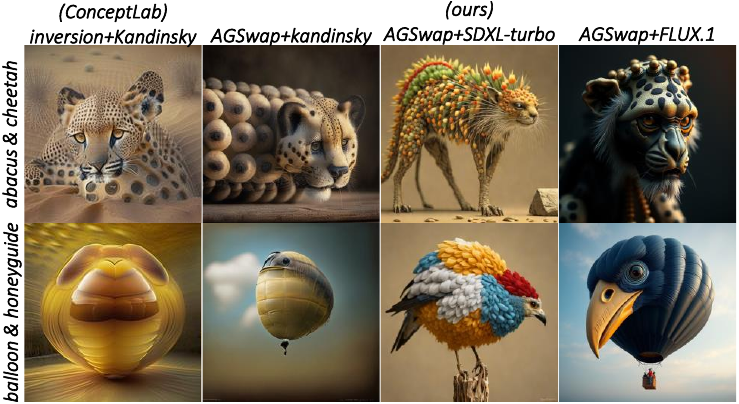}
\caption{Ablation of replacing the backbone with Kandinsky 2.2 and FLUX.1.}
\label{fig:ab_baseline}
\end{figure}

\begin{figure}[t]
\centering
\includegraphics[width=\linewidth]{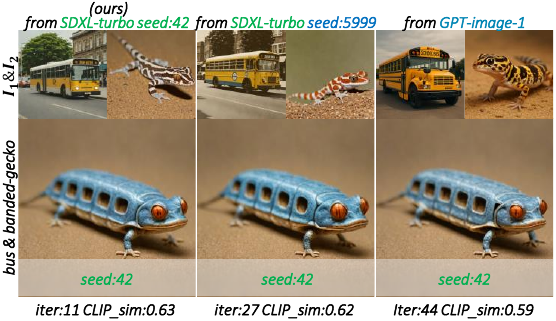}
\caption{Ablation of different images $I_1$ and $I_2$.}
\label{fig:diffimages}
\end{figure}

\textbf{Ablation Study.} \textit{Impact of Reference Image Variations $I_1$ and $I_2$.} We first analyze the role of reference images $I_1$ and $I_2$, generated using category labels $T_1$ and $T_2$ with the same random seed as the current iteration. To isolate their influence, we compare \textit{seed variations} where $I_1/I_2$ are regenerated with different random seeds, and \textit{model variations}, where $I_1/I_2$ are synthesized by GPT-image-1 (same prompts/seed). Fig. \ref{fig:diffimages} shows that while the final output remains stable across variations, convergence speed and similarity scores are affected, suggesting that reference image selection primarily impacts optimization dynamics rather than output quality.

\textit{Robustness to Backbone Model Choice.} To assess generalizability, we evaluate AGSwap on multiple generators by replacing SDXL-Turbo with Kandinsky 2.2 \cite{kandinsky} and FLUX.1-Schnell \cite{flux}, as shown in Fig. \ref{fig:ab_baseline}. On the same backbone, AGSwap consistently surpasses ConceptLab with more coherent single-object fusions; across backbones—including a weaker Kandinsky baseline and a stronger FLUX variant—AGSwap transfers without retraining and maintains reliable quality, demonstrating both model-agnostic efficacy and practical adaptability. 
\begin{figure}[t]
\centering
\includegraphics[width=\linewidth]{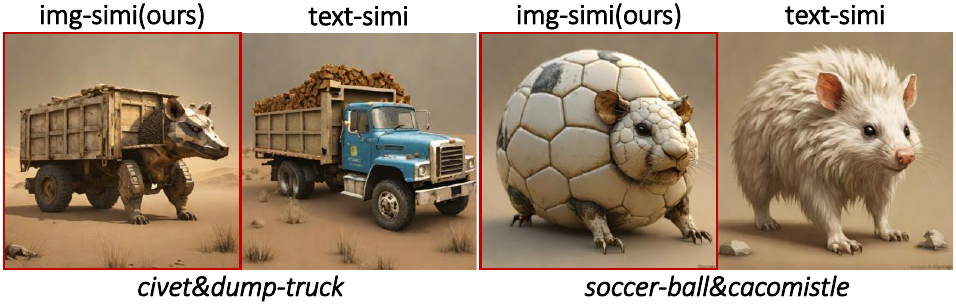}
\caption{Ablation of image- vs. text-based similarities.}
\label{fig:diffimage-text}
\end{figure}

\textit{Image- vs. Text-Based Similarity Metrics.} We next replace $I_1/I_2$ with their text prompts, computing text-image similarity instead of image-image similarity. Fig. \ref{fig:diffimage-text} reveals degraded fusion performance, confirming that text-based alignment fails to capture nuanced visual relationships critical for high-quality synthesis..

\section{Conclusion}
We presented Adaptive Group Swapping (AGSwap), a simple yet highly effective method for synthesizing creative objects from cross-category text pairs. AGSwap integrates two core components: group-wise embedding swapping merges semantic attributes from distinct concepts through feature-level manipulation, and adaptive group updating is a dynamic optimization mechanism guided by a balance evaluation score to ensure coherent synthesis. To support this work, we introduce Cross-category Object Fusion (COF), a large-scale, hierarchically structured dataset derived from ImageNet-1K and WordNet. Extensive experiments demonstrate that AGSwap outperforms existing combinational approaches and even surpasses GPT-image-1 across both simple and complex prompt settings.

\clearpage
\newpage

\clearpage
\newpage
\bibliographystyle{ACM-Reference-Format}
\bibliography{sample-sigconf}

\clearpage
\newpage
\appendix

This supplementary material provides additional details of our paper. We first discuss the limitations of our method in Appendix \ref{sec:sup_limi}, along with several failure cases. In Appendix \ref{sec:cofdatasetappendix}, we present the removed superclass names that were filtered during the initial screening of the COF dataset, as well as the complete version of the dataset’s superclasses and subclasses. Appendix \ref{sec:sup_stu} provides a detailed breakdown of the final vote counts for each question in the user study. Lastly, in Appendix \ref{sec:sup_more}, we show additional results of our method, including: (1) more fusion examples, (2) stylized fusion outputs, and (3) fusion results with manually specified bias, further demonstrating the potential of our approach.

\section{Limitation}
\label{sec:sup_limi}
\begin{figure}[h]
\centering
\includegraphics[width=0.97\linewidth]{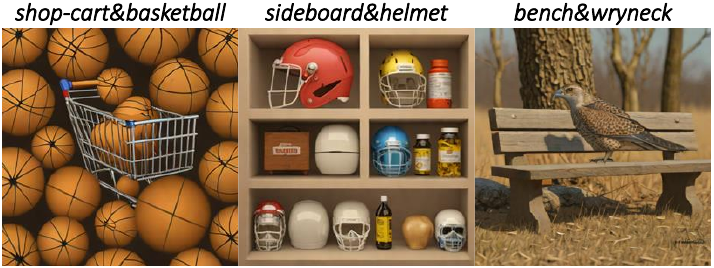}
\caption{Failure Examples} 
\label{fig:sup_fail}
\end{figure}

 While our AGSwap achieves strong qualitative and quantitative results, two key challenges remain. \textit{Evaluation of Creativity:} No direct metrics exist to assess the creativity of fused outputs, as reflected in our quantitative analysis. \textit{Failure Cases:} Certain category combinations still lead to unsuccessful fusions; additional examples are provided in Fig. \ref{fig:sup_fail}.
 
\section{COF dataset}
\label{sec:cofdatasetappendix}

Table \ref{table:COF_all} presents the detailed contents of the COF dataset. And 390 deleted category in $|S_{\text{can}}|$ of Eq. \eqref{eq:scan} are shown in Table \ref{tab:delet_cof}

\begin{figure}[t]
\centering
\includegraphics[width=0.97\linewidth]{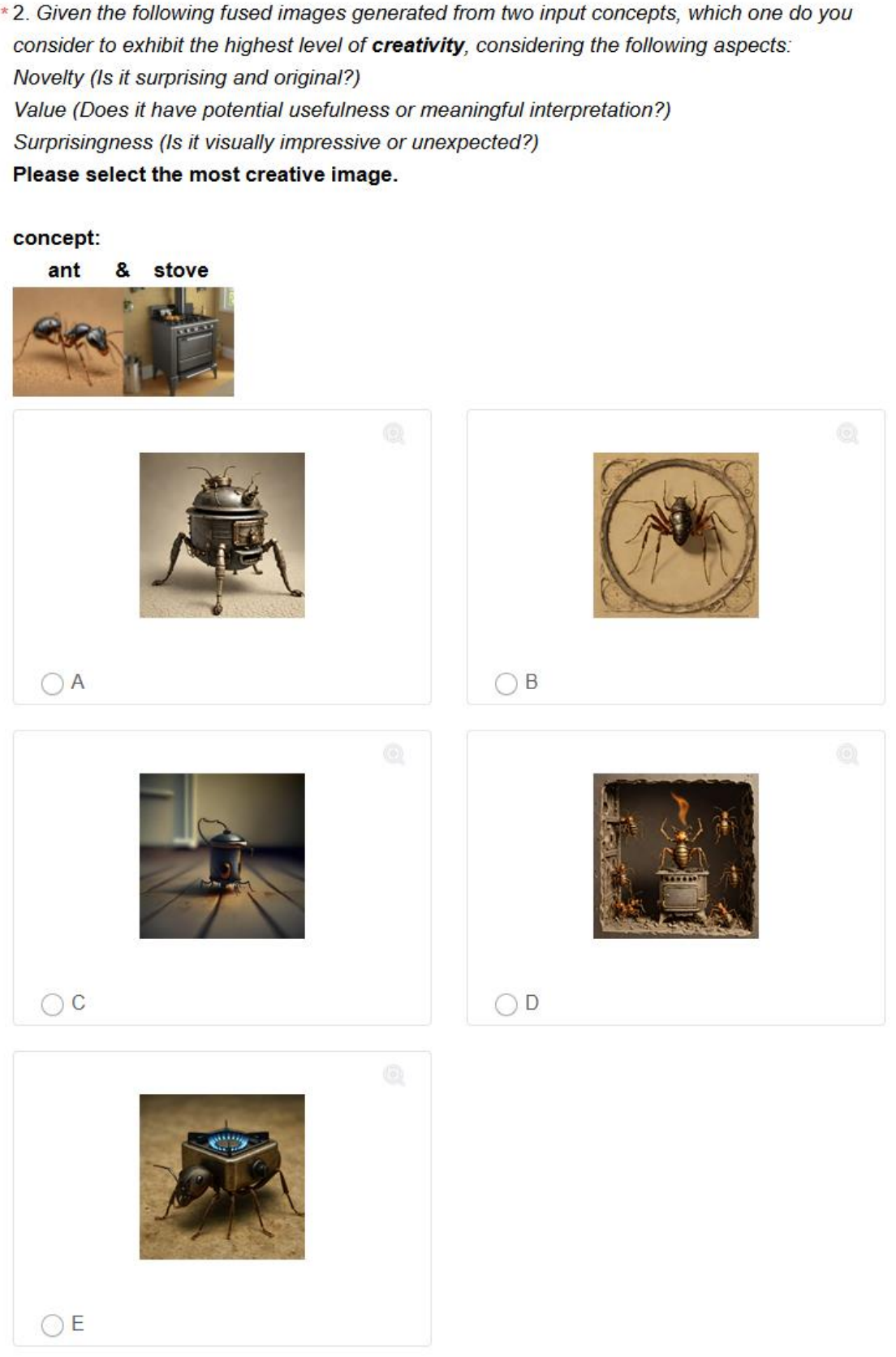}
\caption{More Visual Results} 
\label{fig:sup_usr}
\end{figure}

\section{User Study}
\label{sec:sup_stu}
We provide a more detailed description of the user study. The questions are illustrated in Fig. \ref{fig:sup_usr}, and the specific vote counts for each question are reported in Table \ref{tab:sup_votes1} and Table \ref{tab:sup_votes2}. As shown in the table, our method consistently received the highest number of votes, demonstrating its advantage in perceived creativity. In the complex prompt voting results, the images generated using prompts derived from our synthesized images received more votes than those generated directly from GPT-4o prompts. This confirms that our method can effectively guide generative models to produce more creative outputs.

\section{More Results}
\label{sec:sup_more}
\subsection{More visual Results}
\begin{figure*}[t]
\centering
\includegraphics[width=0.9\linewidth]{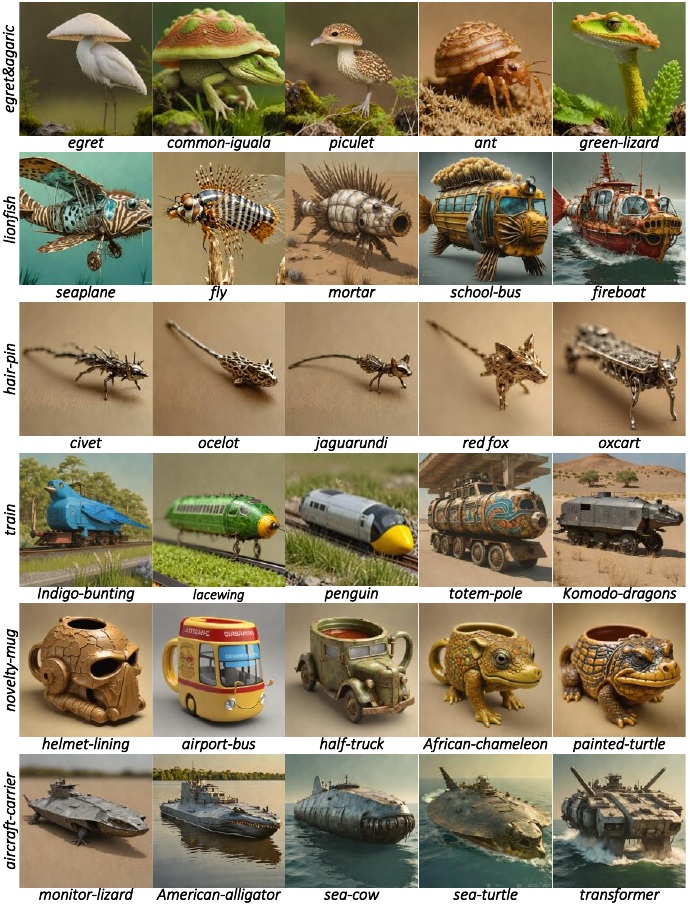}
\caption{More Visual Results} 
\label{fig:sup_more_res}
\end{figure*}

More visual results are shown in Fig. \ref{fig:sup_more_res}

\subsection{Stylized Results}
\begin{figure*}[t]
\centering
\includegraphics[width=0.9\linewidth]{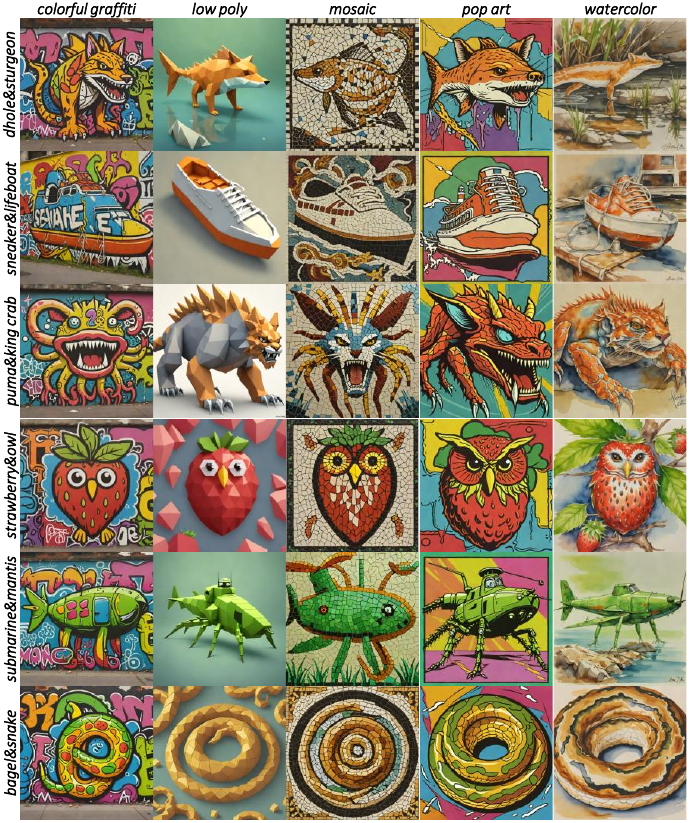}
\caption{More Visual Results} 
\label{fig:sup_style}
\end{figure*}

Fig. \ref{fig:sup_style} presents the results of our method under stylized prompts. We adopted five distinct styles: colorful graffiti, lowpoly, mosaic, pop art, and watercolor. The input prompts were modified to follow the format "A [style] of [word1]" and "A [style] of [word2]". As shown in the figure, our method is still able to generate creative and coherent fusion results under stylized descriptions, demonstrating its effectiveness across different visual styles.

\subsection{Manually Controlled Bias in Fusion Results}
\begin{figure*}[t]
\centering
\includegraphics[width=0.9\linewidth]{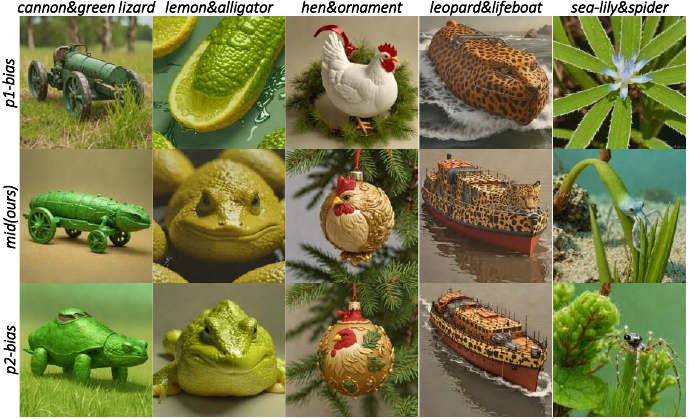}
\caption{Biased fusion results} 
\label{fig:sup_bias}
\end{figure*}

Fig. \ref{fig:sup_bias} presents the results under manually specified bias. We modify Eq. \eqref{eq:score} to the following form:
\begin{align}
    s = (\alpha_{\text{left}} \cdot s_\beta + d_1) - (\alpha_{\text{right}} \cdot s_\beta + d_2),
\end{align}
where $\alpha_{\text{left}}$ and $\alpha_{\text{right}}$ are manually assigned. When the user specifies a right-side bias, we set $\alpha_{\text{left}} = 1$ and $\alpha_{\text{right}} = 0$. In this case, the final image will have a higher similarity to $I_1$ than to $I_2$, with a difference equal to $s_\beta$. We set $s_\beta = 0.05$ to ensure that the generated image exhibits a biased yet still coherent fusion.

\begin{table*}[htbp]
\centering
\setlength{\tabcolsep}{7pt}
\renewcommand{\arraystretch}{1.1}
\caption{User study with combinational methods.}
\resizebox{0.77\linewidth}{!}{
\begin{tabular}{c||c|c|c|c|c}
\Xhline{1.2pt} 
\diagbox{text-pair}{Method} & A(Ours) & B(BASS) & C(ConceptLab) & D(SDXL-turbo) & E(GPT-image-1)\\ 
\hline
bee\&broccoli  & 50(68.49\%) & 1(1.37\%) & 15(20.55\%) & 6(8.22\%) & 1(1.37\%) \\
ant\&stove  & 53(72.6\%) & 3(4.11\%) & 5(6.85\%) & 5(6.85\%) & 7(9.59\%) \\
cannon\&lionfish  & 44(60.27\%) & 2(2.74\%) & 17(23.29\%) & 4(5.48\%) & 6(8.22\%) \\
coho\&sneaker  & 68(93.15\%) & 0(0\%) & 1(1.37\%) & 4(5.48\%) & 0(0\%) \\
hawk\&basket star  & 54(73.97\%) & 11(15.07\%) & 3(4.11\%) & 1(1.37\%) & 4(5.48\%) \\
mantis\&bike  & 47(64.38\%) & 20(27.40\%) & 2(2.74\%) & 3(4.11\%) & 1(1.37\%) \\
\Xhline{1.2pt}
\end{tabular}}
\label{tab:sup_votes1}
\end{table*}
\begin{table*}[htbp]
\centering
\setlength{\tabcolsep}{7pt}
\renewcommand{\arraystretch}{1.1}
\caption{User study with prompt engineering.}
\resizebox{0.77\linewidth}{!}{
\begin{tabular}{c||c|c|c|c|c}
\Xhline{1.2pt} 
\multirow{2}{*}{\diagbox{text-pair}{Method}} & \multirow{2}{*}{A(Ours)} & \multicolumn{2}{c|}{GPT-4o-Generated Prompts} & \multicolumn{2}{c}{Image-Derived Prompts}\\
\cline{3-6}
~ & ~ & B(SDXL-turbo) & C(GPT-image-1) & D(SDXL-turbo) & E(GPT-image-1)\\ 
\hline
bittern\&perfume  & 46(63.01\%) & 2(2.74\%) & 0(0\%) & 24(32.88\%) & 1(1.37\%) \\
titi\&loggerhead  & 30(41.10\%) & 4(5.48\%) & 18(24.66\%) & 6(8.22\%) & 15(20.55\%) \\
great-grey-owl\&drone  & 44(60.27\%) & 3(4.11\%) & 0(0\%) & 4(5.48\%) & 22(30.14\%) \\
bubble\&black-widow  & 37(50.68\%) & 2(2.74\%) & 8(10.96\%) & 21(28.77\%) & 5(6.85\%) \\
\Xhline{1.2pt}
\end{tabular}}
\label{tab:sup_votes2}
\end{table*}
\begin{table*}[t]
\centering
\scriptsize
\setlength{\tabcolsep}{4pt}
\renewcommand{\arraystretch}{0.97}
\caption{List of All deleted Superclasses in Our Dataset (Sorted by Word Length)}
\resizebox{0.97\linewidth}{!}{%
\begin{tabular}{c|c|c|c|c|c|c|c|c|c|c|c}
\Xhline{1.2pt}
bar & mask & finch & tunic & rocket & dynasty & crucifer & framework & ballplayer & wading bird & saltwater fish & cleaning implement \\
bat & newt & fruit & wheel & rodent & factory & cyprinid & gastropod & bedclothes & white goods & selfish person & even-toed ungulate \\
bin & oven & glove & action & runner & firearm & dwelling & great ape & body armor & wolf spider & skilled worker & homopterous insect \\
bun & pipe & heron & agamid & salmon & garment & european & greyhound & collection & bird of prey & anseriform bird & medical instrument \\
cap & post & hound & agaric & screen & globule & farmhand & hairpiece & concoction & broadcasting & armored vehicle & natural depression \\
car & rack & knife & bridge & series & helping & fastener & hamburger & court game & containerful & converging lens & optical instrument \\
dip & robe & lemur & canopy & shaker & hosiery & kangaroo & hand tool & duplicator & corvine bird & cooking utensil & protective garment \\
fly & room & lever & cattle & sheath & jewelry & memorial & harvester & fruit tree & domestic cat & decision making & reproductive organ \\
fur & seed & light & cereal & shrine & leporid & minibike & hindrance & hand glass & farm machine & four-minute man & orthopterous insect \\
hat & shed & maker & citrus & spring & mansion & mountain & historian & lesser ape & flag officer & horny structure & protective covering \\
jar & shoe & melon & coffee & stereo & odonate & pinscher & implement & projectile & illustration & spaghetti sauce & religious residence \\
pan & shop & meter & column & stupid & oilseed & platform & marsupial & prominence & shepherd dog & tracked vehicle & stringed instrument \\
pen & sink & noise & cuckoo & system & opening & position & monotreme & receptacle & siphonophore & wind instrument & free-reed instrument \\
rod & sled & organ & driver & tablet & pointer & pullover & motorboat & scorpaenid & teiid lizard & belgian sheepdog & hymenopterous insect \\
area & soup & penis & equine & thrush & pudding & religion & nonworker & small boat & undergarment & breathing device & measuring instrument \\
base & trap & pouch & fabric & turner & russian & sailboat & nymphalid & spectacles & bottle opener & chest of drawers & pelecaniform seabird \\
bill & tube & punch & filter & turtle & sea cow & sea boat & phalanger & supervisor & chief justice & coraciiform bird & surveying instrument \\
boat & wolf & river & flower & vessel & shelter & sled dog & plaything & armor plate & concave shape & dipterous insect & percussion instrument \\
book & worm & scarf & fodder & weight & spaniel & standard & restraint & ceratopsian & conveyer belt & metallic element & double-reed instrument \\
cart & atoll & shade & fungus & worker & striker & subshrub & sandglass & chordophone & exhibitionist & musteline mammal & heavier-than-air craft \\
case & berry & shore & ganoid & writer & support & swimsuit & sandpiper & compartment & farm building & national capital & lighter-than-air craft \\
city & board & skirt & grouse & admiral & surface & titmouse & shellfish & evil spirit & head of state & new world monkey & self-propelled vehicle \\
dish & brass & smell & hairdo & athlete & sweater & watchdog & shorebird & freight car & motor vehicle & old world monkey & correctional institution \\
film & bunch & snipe & helmet & barrier & terrier & wild dog & strongbox & neuropteron & piciform bird & place of worship & malacostracan crustacean \\
fish & catch & sound & island & big cat & toy dog & wildfowl & supporter & nightingale & visual signal & rescue equipment & bowed stringed instrument \\
fuel & check & spitz & keeper & browser & trouser & woodwind & timepiece & participant & american state & volatile storage & electromagnetic radiation \\
gate & cloak & staff & makeup & chicken & wildcat & artillery & true frog & rattlesnake & apodiform bird & copper-base alloy & electro-acoustic transducer \\
gown & corgi & stick & marker & closure & airplane & ball game & vegetable & round shape & burial chamber & kitchen appliance &  \\
herb & crack & sweet & marrow & concept & american & butterfly & wild goat & sennenhunde & chair of state & natural elevation &  \\
knot & cream & swine & medium & control & antelope & combatant & wolfhound & sense organ & golf equipment & portable computer &  \\
lamp & dress & table & parrot & cracker & arachnid & emptiness & adventurer & submersible & high-angle gun & structural member &  \\
lock & fairy & thief & pistol & cushion & baby bed & enclosure & ambystomid & time period & natural object & unpleasant person &  \\
lory & fiber & timer & record & decline & building & food fish & applicator & toy spaniel & optical device & angiospermous tree &  \\
\Xhline{1.2pt}
\end{tabular}}
\label{tab:delet_cof}
\end{table*}


\clearpage
\newpage
\onecolumn

\setlength{\tabcolsep}{5pt}
\renewcommand{\arraystretch}{1.1}
\centering

\begin{longtable}{l|p{14cm}}
\caption{Details of the COF Dataset (Continued on Next Page)} \\
\hline
\textbf{Superclass} & \textbf{Subclasses} \\
\hline
\endfirsthead

\caption*{Details of the COF Dataset (Continued)} \\
\hline
\textbf{Superclass} & \textbf{Subclasses} \\
\hline
\endhead

\midrule
\multicolumn{2}{r}{\textit{Continued on Next Page}} \\
\endfoot

\bottomrule
\endlastfoot

canine & red wolf, hyena, red fox, grey fox, white wolf, dhole, dingo, Arctic fox, African hunting dog, fennec fox \\
\hline
dog & English setter, Siberian husky, Australian terrier, English springer, malamute, Great Dane, Walker hound, Welsh springer spaniel, Scottish deerhound, Weimaraner \\
\hline
marine mammals & grey whale, sea lion, killer whale, dugong, sea cow, walrus, bottlenose dolphin, harbor seal, Florida manatee, blue whale \\
\hline
procynoid & lesser panda, giant panda, ringtail, raccoon, Coati, kinkajou, cacomistle, mountain coatimundi, eastern mountain coatimundi, olingo \\
\hline
cat & Egyptian cat, Persian cat, cougar, tiger cat, Siamese cat, tabby, lynx, bobcat, serval, caracal \\
\hline
bovid & Alpine Ibex, Gazelle, Bighorn, Hartebeest, Ox, Water Buffalo, Bison, Kudu, Saiga Antelope, Yak \\
\hline
rodnet & porcupine, fox squirrel, hamster, marmot, mouse, rat, beaver, chinchilla, gerbil, capybara \\
\hline
person & groom, scuba diver, ballplayer, pirate, pitcher, suit, loafer, dumbbell, punching bag, doormat \\
\hline
musteline & mink, black footed ferret, polecat, otter, stoat, ermine, marten, wolverine, honey badger, sable \\
\hline
pachyderm & African elephant, Indian elephant, white rhinoceros, black rhinoceros, Sumatran rhinoceros, Javan rhinoceros, greater one horned rhinoceros, tapir, hippopotamus, warthog \\
\hline
large felines & leopard, snow leopard, cheetah, lion, tiger, puma, lynx, jaguar, jaguarundi, ocelot \\
\hline
viverrine & meerkat, mongoose, civet, genet, linsang, banded mongoose, dwarf mongoose, slender mongoose, white tailed mongoose, marsh mongoose \\
\hline
primate & titi, colobus, guenon, indri, orangutan, squirrel monkey, chimpanzee, proboscis monkey, gorilla, spider monkey \\
\hline
edentate & three toed sloth, armadillo, tamandua, silky anteater, giant anteater, two toed sloth, brown throated sloth, pale throated sloth, nine banded armadillo, southern tamandua \\
\hline
\hline
ungulate & sorrel, wild boar, zebra, warthog, Arabian camel, hippopotamus, llama, horse, deer, giraffe \\
\hline
bear & brown bear, ice bear, black bear, sloth bear, spectacled bear, grizzly bear, American black bear, Asiatic black bear, sun bear, Kodiak bear \\
\hline
lagomorph & hare, wood rabbit, domestic rabbit, snowshoe hare, European hare, mountain hare, Arctic hare, pika, cottontail rabbit, volcano rabbit \\
\hline
metatherian & koala, wallaby, wombat, kangaroo, opossum, Tasmanian devil, quokka, sugar glider, bandicoot, Tasmanian tiger \\
\hline
other mammal & tusker, bat, shrew, hedgehog, mole, pangolin, aardvark, hyrax, tenrec, solenodon \\
\hline
prototherian & echidna, platypus, short beaked echidna, eastern long beaked echidna, western long beaked echidna, Attenborough's long beaked echidna, mountain echidna, New Guinea echidna, Tasmanian echidna, Australian echidna \\
\hline
weapon & revolver, missile, bow, assault rifle, rifle, projectile, pistol, shotgun, sword, knife \\
\hline
covering & umbrella, shower curtain, manhole cover, book jacket, ski mask, bottlecap, tile roof, shield, scabbard, lampshade \\
\hline
sailing vessel & schooner, yawl, catamaran, trimaran, sloop, ketch, gaff rig, cutter, brig, lugger \\
\hline
ball & soccer ball, golf ball, ping pong ball, rugby ball, volleyball, tennis ball, bubble, basketball, billiard ball, bowling ball \\
\hline
musical instrument & accordion, grand piano, upright, chime, gong, maraca, marimba, steel drum, banjo, cello \\
\hline
other insect & ant, fly, bee, grasshopper, cricket, walking stick, cicada, leafhopper, lacewing, dragonfly \\
\hline
echinoderm & starfish, sea urchin, sea cucumber, brittle star, feather star, sand dollar, sea lily, basket star, sea daisy, crinoid \\
\hline
mollusk & chambered nautilus, conch, snail, sea slug, chiton, oyster, mussel, clam, octopus, squid \\
\hline
machine & laptop, shovel, chain saw, abacus, cash machine, slide rule, desktop computer, hand held computer, notebook, web site \\
\hline
plant part & strawberry, Granny Smith, orange, lemon, fig, pineapple, banana, jackfruit, custard apple, pomegranate \\
\hline
aircraft & airliner, airship, balloon, plane, helicopter, glider, fighter plane, bomber plane, drone, seaplane \\
\hline
other vehicle & warplane, steamroller, half track, bobsled, dogsled, bumper car, motorcycle, train, thruster, tram \\
\hline
other equipment & space shuttle, reel, castle, photocopier, crossword puzzle, Polaroid camera, parachute, jigsaw puzzle, reflex camera, telescope \\
\hline
other craft & Fireboat, Gondola, Speedboat, Lifeboat, Canoe, Container Ship, Wreck, Tugboat, basketry, Sailboat \\
\hline
food & pizza, coho, confectionery, mashed potato, bagel, hot pot, trifle, menu, meat loaf, guacamole \\
\hline
abstraction & pattern, texture, symmetry, geometry, gradient, noise, fractal, silhouette, shadow, reflection \\
\hline
wheeled vehicle & tank, freight car, passenger car, barrow, shopping cart, motor scooter, tricycle, horse cart, jinrikisha, oxcart \\
\hline
bicycle & bicycle built for two, mountain bike, unicycle, road bike, hybrid bike, cruiser bike, folding bike, electric bike, tandem bike, cargo bike \\
\hline
self propelled vehicle & forklift, electric locomotive, steam locomotive, ambulance, beach wagon, cab, convertible, jeep, limousine, minivan \\
\hline
amphibian & Chinese Giant Salamander, Golden Poison Dart Frog, South American Horned Frog, Japanese Giant Salamander, Caecilian, Coquí Frog, Hellbender, Surinam Toad, Tiger Salamander, Glass Frog \\
\hline
snake & racer, thunder snake, ringneck snake, hognose snake, green snake, king snake, garter snake, water snake, vine snake, night snake \\
\hline
sports equipment & golfcart, iron, croquet ball, balance beam, baseball, basketball, barbell, horizontal bar, parallel bars, tennis racket \\
\hline
truck & fire engine, garbage truck, tow truck, trailer truck, moving van, police van, tractor, pickup truck, dump truck, cement mixer truck \\
\hline
structure & mobile home, triumphal arch, patio, steel arch bridge, suspension bridge, viaduct, barn, greenhouse, palace, monastery \\
\hline
container & cradle, ashcan, thimble, cocktail shaker, tub, tray, Petri dish, bucket, soup bowl, mortar \\
\hline
furniture & crib, four poster, bookcase, china cabinet, medicine chest, chiffonier, table lamp, file, park bench, barber chair \\
\hline
body part & hip, ear, organ, hammer, cock, cannon, nail, muzzle, dock, plate \\
\hline
teleost fish & drum, eel, rock beauty, anemone fish, lionfish, puffer, sturgeon, gar, trout, salmon \\
\hline
glass & flute, beer glass, goblet, wine glass, shot glass, liqueur glass, tumbler, carafe, decanter, pitcher \\
\hline
other plant & daisy, yellow lady's slipper, pinwheel, torch, lumbermill, bell pepper, head cabbage, broccoli, cauliflower, zucchini \\
\hline
geological formation & cliff, valley, alp, volcano, promontory, coral reef, lakeside, seashore, geyser, canyon \\
\hline
tool & hatchet, cleaver, letter opener, power drill, lawn mower, corkscrew, can opener, screwdriver, plow, pick \\
\hline
game fowl & hen, black grouse, ptarmigan, ruffed grouse, prairie chicken, peacock, quail, partridge, pheasant, turkey \\
\hline
bird & brambling, goldfinch, house finch, junco, indigo bunting, robin, bulbul, chickadee, water ouzel, African grey \\
\hline
raptors & kite, bald eagle, great grey owl, hawk, falcon, owl, vulture, harrier, eagle, kestrel \\
\hline
woodpeckers & jacamar, toucan, woodpecker, barbet, honeyguide, puffbird, piculet, wryneck, sapsucker, flicker \\
\hline
waterfowl & red breasted merganser, black swan, white stork, black stork, spoonbill, flamingo, American egret, little blue heron, bittern, limpkin \\
\hline
penguins & king penguin, emperor penguin, gentoo penguin, Adélie penguin, chinstrap penguin, rockhopper penguin, macaroni penguin, little penguin, yellow eyed penguin, fiordland penguin \\
\hline
shark & great white shark, tiger shark, thresher, hammerhead shark, bull shark, lemon shark, nurse shark, whale shark, basking shark, blue shark \\
\hline
ray & electric ray, stingray, manta ray, eagle ray, cownose ray, butterfly ray, devil ray, skate, guitarfish, sawfish \\
\hline
cypriniform fish & tench, goldfish, carp, minnow, loach, barb, dace, chub, bitterling, rudd \\
\hline
anapsid & leatherback turtle, mud turtle, terrapin, box turtle, sea turtle, tortoise, snapping turtle, painted turtle, red eared slider, softshell turtle \\
\hline
saurian & banded gecko, common iguana, American chameleon, whiptail, agama, frilled lizard, alligator lizard, Gila monster, green lizard, African chameleon \\
\hline
reptile & triceratops, African crocodile, American alligator, tuatara, gharial, Nile crocodile, caiman, komodo dragon, monitor lizard, iguana \\
\hline
other device & paintbrush, hand blower, oxygen mask, snorkel, screen, electric fan, oil filter, strainer, space heater, guillotine \\
\hline
electrical device & loudspeaker, microphone, coil, television, radio, light bulb, battery, transformer, capacitor, resistor \\
\hline
appliance & stove, espresso maker, microwave, rotisserie, waffle iron, vacuum, refrigerator, dishwasher, toaster oven, blender \\
\hline
clock & analog clock, digital clock, wall clock, grandfather clock, wristwatch, alarm clock, cuckoo clock, mantel clock, quartz clock, atomic clock \\
\hline
electronic equipment & joystick, modem, pay phone, radio, oscilloscope, cellular telephone, cassette player, television, dial telephone, tape player \\
\hline
clothing & switch, jean, fur coat, pajama, kimono, knee pad, crash helmet, mask, football helmet, bathing cap \\
\hline
toy & yo-yo, teddy bear, doll, marbles, toy car, toy train set, toy boat, toy airplane, building blocks, slinky \\
\hline
arthropod & trilobite, harvestman, black and gold garden spider, barn spider, garden spider, black widow, tarantula, wolf spider, tick, centipede \\
\hline
crustacean & isopod, Dungeness crab, rock crab, fiddler crab, king crab, American lobster, spiny lobster, crayfish, hermit crab, shrimp \\
\hline
beetle & tiger beetle, ladybug, ground beetle, long horned beetle, leaf beetle, dung beetle, rhinoceros beetle, weevil, firefly, stag beetle \\
\hline
dictyopterous insect & australian cockroach, american cockroach, german cockroach, giant cockroach, mantis, praying mantis, Termite, oriental cockroach, smokybrown cockroach, drywood termite \\
\hline
coelenterate & jellyfish, sea anemone, brain coral, sea fan, Portuguese Man o' War, coral, sea whip, soft coral, hard coral, sea pen \\
\hline
invertebrate & flatworm, nematode, snail, spider, peanut worm, octopus, squid, clam, oyster, mussel \\
\hline
pot & Dutch oven, caldron, coffeepot, teapot, flower pot, soup pot, stock pot, sauce pot, clay pot, pressure cooker \\
\hline
paint & oil paint, acrylic paint, watercolor paint, gouache paint, tempera paint, latex paint, enamel paint, spray paint, primer, varnish \\
\hline
decoration & totem pole, statue, painting, vase, sculpture, tapestry, chandelier, wall hanging, ornament, wreath \\
\hline
fungus & mushroom, hen of the woods, earthstar, coral fungus, stinkhorn, agaric, gyromitra, bolete, truffle, morel \\
\hline
box & carton, safe, pencil box, crate, mailbox, jewelry box, tool box, lunch box, storage box, gift box \\
\hline
piece of cloth & handkerchief, dishrag, paper towel, bib, napkin, tablecloth, curtain, towel, bed sheet, pillowcase \\
\hline
footwear & sandal, running shoe, maillot, Christmas stocking, cowboy boot, clog, sock, sneaker, loafer, flip flop \\
\hline
paddle & table tennis paddle, ping-pong paddle, canoe paddle, kayak paddle, oar, stand-up paddle, rafting paddle, dragon boat paddle, outrigger paddle, pickleball paddle \\
\hline
adornment & necklace, bracelet, earring, ring, brooch, pendant, anklet, cufflink, tiara, hair pin \\
\hline
rug & prayer rug, Persian rug, area rug, carpet, kilim, shag rug, oriental rug, braided rug, throw rug, runner rug \\
\hline
bottle & whiskey jug, beer bottle, pop bottle, water jug, wine bottle, pill bottle, water bottle, flask, ink bottle, specimen bottle \\
\hline
reservoir & water tower, rain barrel, dam, lake, pond, reservoir tank, cistern, water storage, irrigation pond, flood control basin \\
\hline
toiletry & face powder, sunscreen, lipstick, perfume, lotion, hair spray, shampoo, soap, toothbrush, deodorant \\
\hline
bag & mailbag, backpack, envelope, purse, sleeping bag, plastic bag, suitcase, tote bag, grocery bag, duffel bag \\
\hline
gear & drilling platform, carpenter's kit, fishing gear, camping gear, ski gear, diving gear, hiking gear, climbing gear, hunting gear, photography gear \\
\hline
bus & trolleybus, minibus, school bus, city bus, coach bus, double decker bus, articulated bus, shuttle bus, tour bus, airport bus \\
\hline
instrumentality & bullet train, maze, subway train, ferry boat, cable car, monorail, tramway, escalator, conveyor belt, pipeline \\
\hline
padding & pillow, cushion, mattress, foam pad, seat cushion, backrest pad, knee pad, elbow pad, shoulder pad, helmet lining \\
\hline
print media & comic book, magazine, newspaper, book, brochure, catalog, pamphlet, poster, flyer, newsletter \\
\hline
mug & coffee mug, tea mug, travel mug, ceramic mug, insulated mug, glass mug, plastic mug, personalized mug, novelty mug, souvenir mug \\

\label{table:COF_all}
\end{longtable}

\twocolumn

\end{document}